\documentclass{article}

     \PassOptionsToPackage{numbers, compress}{natbib}

     \usepackage[final]{neurips_2020_ml4ps}

\usepackage[utf8]{inputenc} 
\usepackage[T1]{fontenc}    
\usepackage{hyperref}       
\usepackage{url}            
\usepackage{booktabs}       
\usepackage{amsfonts}       
\usepackage{nicefrac}       
\usepackage{microtype}      
\usepackage{subfig}
\usepackage{wrapfig}

\usepackage{amsmath}
\usepackage{amssymb}

\title{Integrable Nonparametric Flows}
\author{David Pfau, Danilo Rezende\\
DeepMind\\
London, UK \\
\texttt{\{pfau, danilor\}@google.com}}

\usepackage{natbib}
\usepackage{graphicx}

\begin{document}

\maketitle

\begin{abstract}
    We introduce a method for reconstructing an infinitesimal normalizing flow given only an infinitesimal change to a (possibly unnormalized) probability distribution. This reverses the conventional task of normalizing flows -- rather than being given samples from a unknown target distribution and learning a flow that approximates the distribution, we are given a perturbation to an initial distribution and aim to reconstruct a flow that would generate samples from the known perturbed distribution. While this is an underdetermined problem, we find that choosing the flow to be an integrable vector field yields a solution closely related to electrostatics, and a solution can be computed by the method of Green's functions. Unlike conventional normalizing flows, this flow can be represented in an entirely nonparametric manner. We validate this derivation on low-dimensional problems, and discuss potential applications to problems in quantum Monte Carlo and machine learning.
\end{abstract}

\section{Introduction}

Consider the case where we are trying to optimize some stochastic objective from samples, where the samples themselves are generated from a distribution that depends on the function we are optimizing. That is, given a parameterized function $f_\theta(\mathbf{x})$ defined over points $\mathbf{x}\in\mathbb{R}^n$, a probability distribution $p\left[f_\theta \right](\mathbf{x})$ that can be derived from $f_\theta$, and an objective functional $\mathcal{L}\left[f_\theta \right](\mathbf{x})$, we are trying to solve:

\begin{equation}
    \min_{\theta} \mathbb{E}_{\mathbf{x}\sim p\left[f_\theta\right]}\left[\mathcal{L}\left[f_\theta\right](\mathbf{x})\right]
\end{equation}
via samples from the empirical distribution. For clarity, we will denote $p\left[f_\theta \right](\mathbf{x})$ by $p_\theta(\mathbf{x})$ from here on out. If $p_\theta(\mathbf{x})$ is some complex distribution that cannot be sampled directly, techniques such as Markov Chain Monte Carlo (MCMC) must be employed, which may be slow to converge to the target distribution.

Problems of this form appear in stochastic variational inference \cite{hoffman2013stochastic}, where the loss is the variational free energy of the approximate posterior, and in policy gradient methods for reinforcement learning \cite{williams1992simple}, where the loss is the long-term cost function and the sampling distribution is the policy. In these cases the distribution $p_\theta$ is usually chosen to be easy to directly sample from, so equilibration is less of an issue. Our motivating example for studying problems of this form is the variational quantum Monte Carlo method in computational physics \cite{foulkes2001quantum}. Here, the function being optimized is the wavefunction $\psi_\theta(\mathbf{x})$, the samples $\mathbf{x}$ represent possible configurations of particles, the probability distribution is proportional to the wavefunction squared $p_\theta(\mathbf{x})\propto |\psi_\theta (\mathbf{x})|^2$, and the objective functional to minimize is the {\em local energy} $\psi^{-1}(\mathbf{x}) \hat{H}\psi(\mathbf{x})$ where $\hat{H}$ is a Hermitian linear operator called the Hamiltonian, and we have dropped the $\theta$ for clarity. The wavefunction typically has a complex form that precludes direct sampling from $p_\theta(\mathbf{x})$, so MCMC methods are employed. Normalizing flows can also be employed instead of MCMC \cite{cranmer2019inferring}, but the computational demands of computing the full Jacobian of the flow make scaling to higher-dimensional problems difficult.

In this case, either a slow learning rate or many MCMC steps between parameter updates must be used for optimizing $\theta$, otherwise the samples may diverge from the target distribution and the optimization may behave pathologically. In many ways this is analogous to the problems encountered in bilevel optimization \cite{colson2007overview} which crop up frequently in machine learning \cite{pfau2016connecting}, where two coupled objectives must be optimized simultaneously. Here instead of two coupled minimization problems, we have a minimization problem coupled to a sampling problem. One way to accelerate learning is to optimize the mixing rate of the MCMC chain -- however, this does not take advantage of our knowledge about how $p_\theta$ is changing.

In principle, if we have samples $\mathbf{x}^1_t, \ldots \mathbf{x}^n_t$ from $p_{\theta_t}$, the target distribution at time $t$, then we should only need to make a small deterministic correction $\mathbf{v}^1_t, \ldots \mathbf{v}^N_t$ to the samples to transform them into samples from the distribution at time $t$+1, assuming the update $\Delta \theta_t = \theta_{t+1} - \theta_t$ is sufficiently small. This should be more efficient at generating samples from the new target density $p\left[f_{\theta_{t+1}}\right]$ than pure MCMC, as it exploits information about the density at the previous time step, rather than only using information about the current time step. This can be seen as an approximate step in a continuous-time normalizing flow such as Neural ODEs \cite{chen2018neural}, where the corrections $\mathbf{v}^1_t, \ldots \mathbf{v}^N_t$ approximate a time-dependent flow field $\mathbf{v}(\mathbf{x}; t)$ that tracks the time evolution of $p_{\theta_t}$. No parametric form for the flow field $\mathbf{v}(\mathbf{x}, t)$ needs to be learned -- only the corrections to individual samples at each time step are needed. In the next section, we will review normalizing flows and discuss how to construct such a set of corrections $\mathbf{v}^1_t, \ldots \mathbf{v}^N_t$ given only knowledge of $p_{\theta_{t}}$ and $p_{\theta_{t+1}}$.

\section{Method}

\paragraph{Normalizing Flows} In conventional normalizing flows \cite{rezende2015variational}, an initial probability distribution $p_0(\mathbf{x})$ with known density that is easy to sample from is transformed deterministically by an invertible function $f$ giving samples $\mathbf{z} = f(\mathbf{x})$ from the distribution $p(\mathbf{z})$. The classic rule for changes to a probability distribution under change of variable give the log probability as 

\begin{equation}
    \mathrm{log}p(\mathbf{z}) = \mathrm{log}p_0(f^{-1}(\mathbf{z})) - \mathrm{log}\left|\mathbf{J}_f(f^{-1}(\mathbf{z}))\right|
    \label{eqn:normalizing_flows}
\end{equation}
where $\mathbf{J}_f$ is the Jacobian of $f$. Normalizing flows estimate the probability distribution $p$ by learning the function $f$. As computing the log determinant scales as $\mathcal{O}(N^3)$ with the number of variables in the most general case, clever choices must be made for the class of functions to learn, such as real non-volume-preserving flows (Real NVP) \cite{dinh2017density}.

\paragraph{Infinitesimal Flows} Another way to reduce the computational overhead of normalizing flows is to use an ordinary differential equation to generate $f$ \cite{chen2018neural}. In this case, the probability distribution changes over a finite time from $p(\mathbf{x}; 0)$ to $p(\mathbf{z}; T)$, where $\mathbf{z}$ is the end point of a curve defined by the ODE $\dot{\mathbf{x}}(t) = \mathbf{v}(\mathbf{x}(t))$, $\mathbf{x}(0) = \mathbf{x}$. For a small time step $dt$, we can approximate $\mathbf{x}(t+dt)$ to first order as $\mathbf{x}(t+dt) = \mathbf{x}(t) + dt \mathbf{v}(\mathbf{x}(t)) + \mathcal{O}(dt^2)$. Plugging this into Eq.~\ref{eqn:normalizing_flows} yields:

\begin{align}
    \mathrm{log}p(\mathbf{x} + dt \mathbf{v}(\mathbf{x}) + \mathcal{O}(dt^2); t+dt) &= \mathrm{log}p(\mathbf{x}; t) - \mathrm{log}\left|\mathbf{J}_{f}(\mathbf{x})\right| \\
    &= \mathrm{log}p(\mathbf{x}; t) - \mathrm{log}|\mathbf{I} + dt \mathbf{J}_\mathbf{v}(\mathbf{x}) + \mathcal{O}(dt^2)|
\end{align}
Taking a Taylor series gives:
\begin{equation}
    \mathrm{log}p(\mathbf{x}; t+dt) + dt \mathbf{v}(\mathbf{x})^T \nabla \mathrm{log}p(\mathbf{x}; t+dt) = \mathrm{log}p(x; t) - dt\mathrm{Tr}(\mathbf{J}_v(\mathbf{x})) + \mathcal{O}(dt^2)
\end{equation}
which, in the limit as $dt\rightarrow 0$, becomes:
\begin{equation}
    \frac{\partial \mathrm{log} p(\mathbf{x}; t)}{\partial t} = -\mathbf{v}(\mathbf{x})^T \nabla \mathrm{log}p(\mathbf{x}; t) - \mathrm{Tr}(\mathbf{J}_\mathbf{v}(\mathbf{x}))  = - \mathbf{v}^T \nabla \mathrm{log}p(\mathbf{x}; t) - \nabla \cdot \mathbf{v}
    \label{eqn:neural_ode}
\end{equation}
after some rearranging of terms. Here $\nabla \cdot$ is the divergence of a vector field, which is just another way of writing the trace of the Jacobian. The right-hand side of this equation is also the trace of the {\em Stein operator} of the distribution $p(\mathbf{x})$ applied to the function $\mathbf{v}(\mathbf{x})$, and plays a critical role in Stein variational gradient descent (SVGD) \cite{liu2016stein}. Switching from the log density to the density (and dropping the $t$ for clarity), we find this expression can be simplified considerably:
\begin{align}
    \frac{1}{p(\mathbf{x})}\frac{\partial p(\mathbf{x})}{\partial t} &= -\mathbf{v}^T\frac{\nabla p(\mathbf{x})}{p(\mathbf{x})} - \nabla \cdot \mathbf{v} \nonumber \\
    \frac{\partial p(\mathbf{x})}{\partial t} &= -\mathbf{v}^T \nabla p(\mathbf{x}) - p(\mathbf{x}) \nabla \cdot \mathbf{v} \nonumber \\
    &= - \nabla \cdot \left(\mathbf{v}(\mathbf{x}) p(\mathbf{x})\right)
    \label{eqn:iinf}
\end{align}
This may also be familiar as the drift term of the Fokker-Planck equation \cite[Eq. 6.48]{kadanoff2000statistical} or the continuity equation for conservation of mass in fluid mechanics. We will denote the change to a probability distribution $\delta p(\mathbf{x})$ rather than $\frac{\partial p(\mathbf{x})}{\partial t}$ throughout the remainder of the paper, as we will not always be in the context of differential equations.

\paragraph{Integrable Nonparametric Flows} 
In neural ODEs, the infinitesimal change to the probability distribution is learned -- all that is given is the final position of the samples, and the field $\mathbf{v}$ is approximated by a neural network. In our case, we assume the change to the probability distribution is known at the start, at least at the position of the samples, and we want to solve Eq.~\ref{eqn:iinf} directly for $\mathbf{v}$. As we are trying to solve for a vector quantity but are only given a scalar, this is underdetermined. However, if we assume that $\mathbf{v}$ is integrable (that it, there exists a scalar function $u(\mathbf{x})$ such that $\mathbf{v}(\mathbf{x})p(\mathbf{x}) = \nabla u(\mathbf{x})$), then Eq.~\ref{eqn:iinf} has the same form as Gauss's law in electromagnetism -- $\mathbf{v}(\mathbf{x}) p(\mathbf{x})$ plays the role of an electric field while $\delta p(\mathbf{x})$ is equivalent to the charge density. This equation admits a general solution of the form
\begin{align*}
\mathbf{v}(\mathbf{x})p(\mathbf{x}) &= \int d\mathbf{z} \delta p(\mathbf{z}) G(\mathbf{x}-\mathbf{z}),
\end{align*}
where $G$ is a convolution kernel known as a ``Green's function" \cite{olenick1986mechanical}, which means that $G$ solves the original equation for a Dirac delta source: $\nabla\cdot G(\mathbf{x}) = \delta(0)$.
In three dimensions, this Green's function takes the familiar form of $\frac{\hat{\mathbf{x}}}{4\pi |\mathbf{x}|^2}$, where $\hat{\mathbf{x}}$ is the unit vector in the direction of $\mathbf{x}$. In $n$ dimensions, this will take the form of $\hat{\mathbf{x}}$ divided by the surface area of the $n-1$ sphere:

\[
G_n(\mathbf{x}) = \frac{\Gamma(\frac{n}{2})\hat{\mathbf{x}}}{2\pi^{n/2}|\mathbf{x}|^{n-1}} = \frac{\Gamma(\frac{n}{2})\mathbf{x}}{2\pi^{n/2}|\mathbf{x}|^n}
\]
Convolving this gives
\begin{align}
\mathbf{v}(\mathbf{x})p(\mathbf{x}) &= \int d\mathbf{z} \delta p(\mathbf{z}) G_n(\mathbf{x}-\mathbf{z}) = \int d\mathbf{z} p(\mathbf{z}) \frac{\delta p(\mathbf{z})}{p(\mathbf{z})} G_n(\mathbf{x}-\mathbf{z}) = \mathbb{E}_{\mathbf{z}\sim p}\left[\frac{\delta p(\mathbf{z})}{p(\mathbf{z})}G_n(\mathbf{x}-\mathbf{z})\right] \nonumber \\
\mathbf{v}(\mathbf{x}) &= \mathbb{E}_{\mathbf{z}\sim p}\left[\frac{\delta p(\mathbf{z})}{p(\mathbf{z})p(\mathbf{x})}G_n(\mathbf{x}-\mathbf{z})\right]
\label{eqn:iinf_expectation}
\end{align}
The Green's function can also be expressed as the gradient of the {\em Coulomb kernel}, that is, $G_n(\mathbf{x}-\mathbf{z}) = \nabla_\mathbf{x} k_n(\mathbf{x}, \mathbf{z})$ where:

\begin{equation}
    k_n(\mathbf{x}, \mathbf{z}) = \frac{\Gamma(\frac{n}{2})}{2(n-2)\pi^{n/2}|\mathbf{x}-\mathbf{z}|^{n-2}} 
\end{equation}
The Coulomb kernel is a natural choice for this problem. Not only is convolution with the gradient of this kernel the only possible solution for Eq.~\ref{eqn:iinf} for which $\mathbf{v}(\mathbf{x})p(\mathbf{x})$ is integrable, but it is also the {\em optimal} choice of kernel to define an energy for aligning distributions via samples following force fields \cite{hochreiter2005optimal}.\footnote{The proof of optimality in \cite{hochreiter2005optimal} can be found at \url{http://www.bioinf.jku.at/publications/2005/ijcnnsupplementary.pdf}}

\paragraph{Unnormalized densities} In cases where MCMC would be used for sampling, we typically only have access to the log of an {\em unnormalized} probability distribution, $\ell(\mathbf{x})$ such that $\mathrm{exp}(\ell(\mathbf{x})) \propto p(\mathbf{x})$. Let us assume we also have an infinitesimal change to the log unnormalized probability $\delta \ell(\mathbf{x})$. Then the expression in Eq.~\ref{eqn:iinf_expectation} simplifies to:
\begin{equation}
    \mathbf{v}(\mathbf{x}) = \mathbb{E}_{\mathbf{z}\sim p}\left[\frac{\delta \ell(\mathbf{z}) - \mathbb{E}_{\mathbf{z}'\sim p}[\delta \ell(\mathbf{z}')]}{p(\mathbf{x})}G_n(\mathbf{x}-\mathbf{z})\right]
    \label{eqn:unnormalized_iinf}
\end{equation}
Unfortunately, a term of the form $p(\mathbf{x})$ persists in the denominator of the expectation. This means that without knowledge of the partition function for the distribution, we can only recover $\mathbf{v}$ up to some constant multiple. In principle, we could use annealed importance sampling \cite{neal2001annealed} to generate an unbiased estimate of the partition function, or we could take the difference between the empirical expectation of $\mathrm{g}(\mathbf{x})$ and an unbiased estimate of the entropy \cite{beirlant1997nonparametric}, which would give an unbiased estimate of the log partition function. We leave it to future work to explore how best to address this.

\section{Experiments}

\begin{figure}
    \centering
    \subfloat[][Density $p(\mathbf{x})$]{\includegraphics[width=0.33\textwidth]{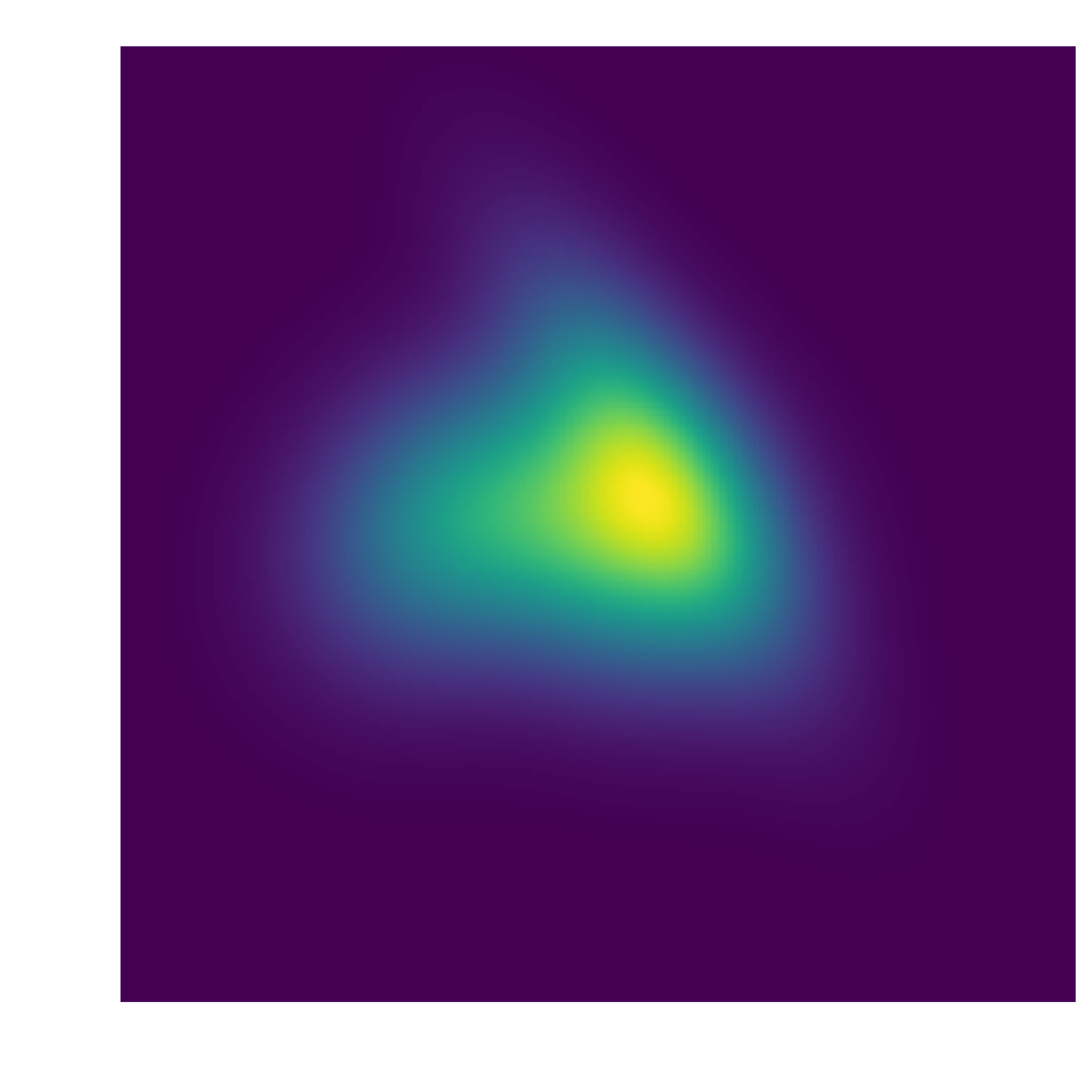}\label{fig:density}}
    \subfloat[][Integrable flow field $\mathbf{v}(\mathbf{x})$, overlaid on perturbation to the density $\delta p(\mathbf{x})$]{\includegraphics[width=0.33\textwidth]{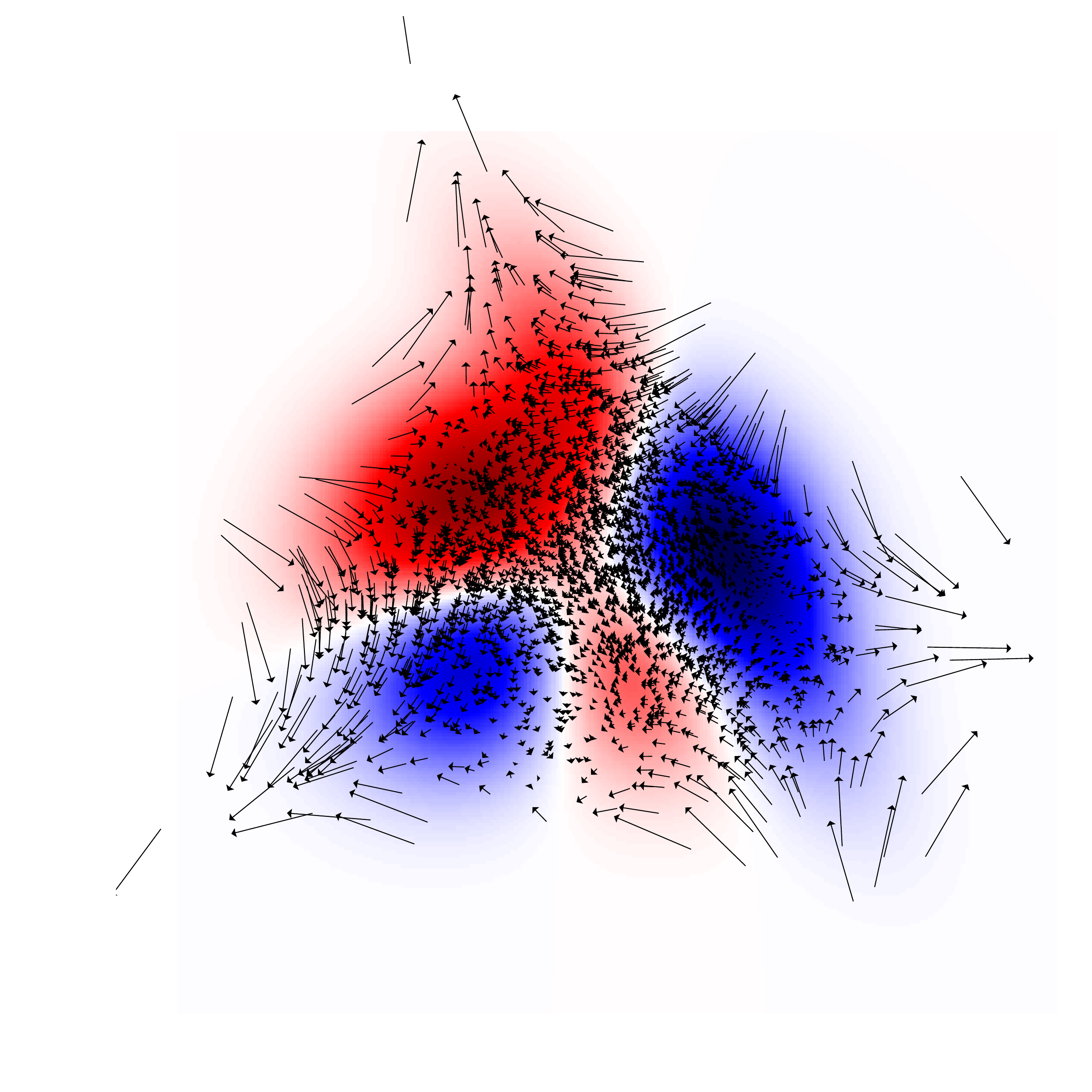}\label{fig:flow}}
    \subfloat[][Kernel density estimate of $\delta p(\mathbf{x})$]{\includegraphics[width=0.33\textwidth]{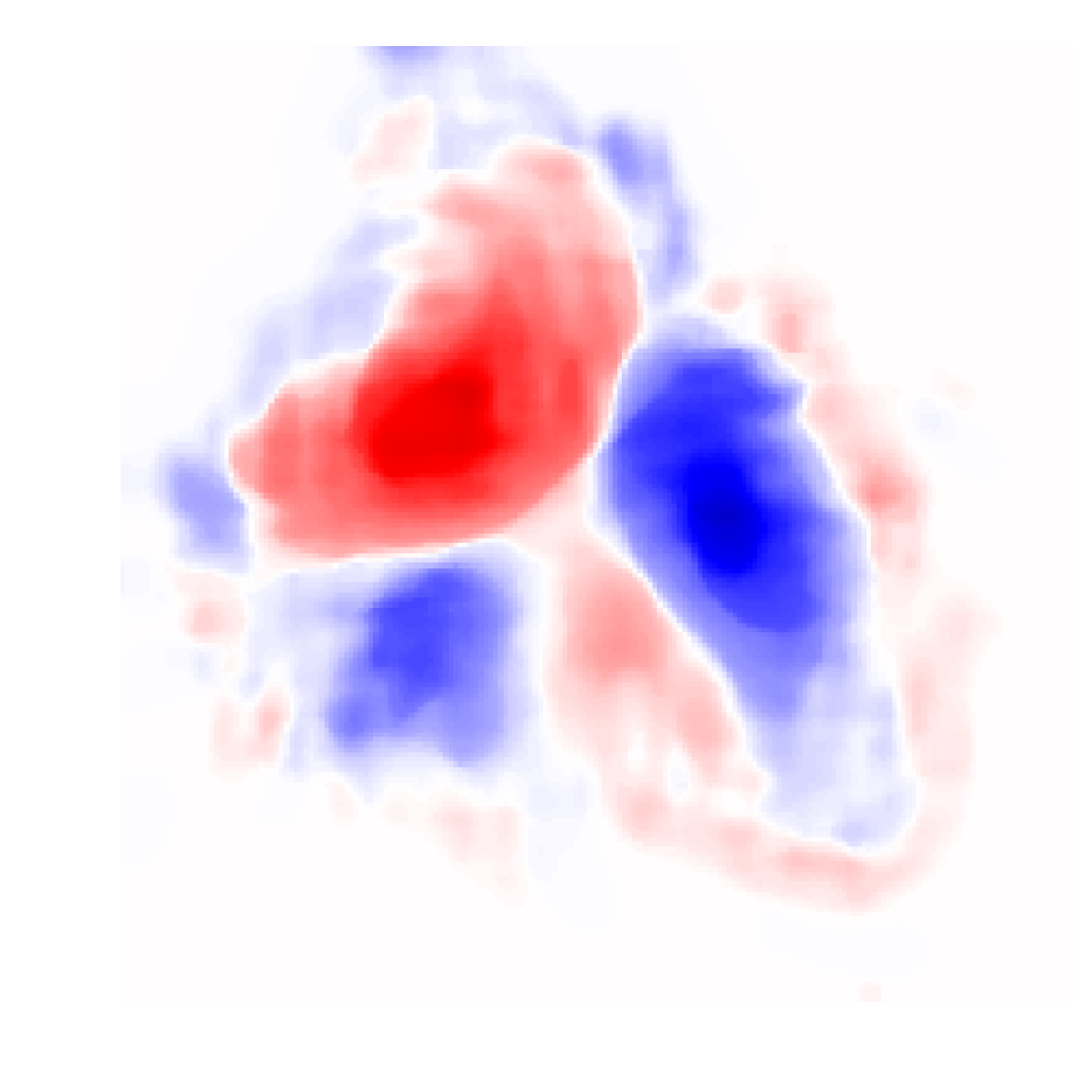}\label{fig:kde}}
    \caption{Visualization of integrable flow field for a mixture of Gaussians.}
\end{figure}

To estimate the quantity in Eq.~\ref{eqn:unnormalized_iinf} empirically, we treat samples from $p(\mathbf{x})$ much like we would treat particles in an electrostatics calculation. Given a set of samples $\mathbf{x}^1,\ldots,\mathbf{x}^N$ from $p(\mathbf{x}) \propto \mathrm{exp}(\ell(\mathbf{x}))$ and a set of perturbations $\delta \ell(\mathbf{x}^1), \ldots, \delta \ell(\mathbf{x}^N)$, the integrable flow is estimated to be:

\begin{equation}
    \mathbf{v}^i \propto \frac{1}{(N-1)\mathrm{exp}(\ell(\mathbf{x}^i))}\sum_{\substack{j=1 \\ j\ne i}}^N \left(\delta \ell(\mathbf{x}^j) - \frac{1}{N} \sum_{k=1}^N \delta \ell(\mathbf{x}^k) \right) G_n(\mathbf{x}^i -  \mathbf{x}^j)
\end{equation}

\begin{wrapfigure}{r}{0.5\textwidth}
    \vspace{-25pt}
    \centering
    \includegraphics[width=0.5\textwidth]{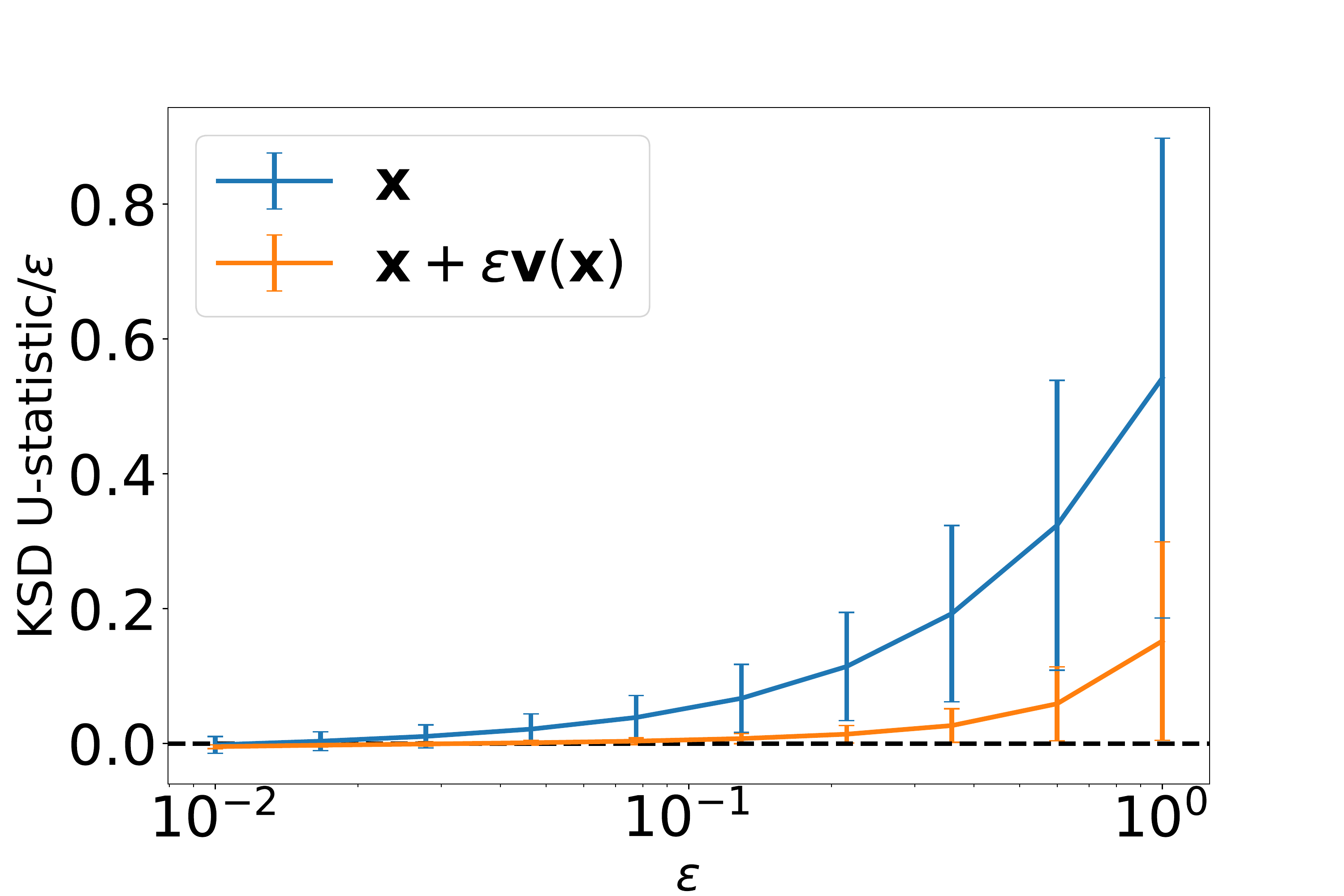}
    \caption{The average empirical kernelized Stein discrepancy between samples $\mathbf{x}^i$ or $\mathbf{x}^i + \epsilon \mathbf{v}^i$ and $p(\mathbf{x}) + \epsilon \delta p(\mathbf{x})$.}
    \label{fig:ksd}
    \vspace{-10pt}
\end{wrapfigure}

As a simple demonstration, we show how this estimate performs on a 2D mixture of Gaussians with 3 components (Fig. 1). We generate 10,000 samples, and then take random perturbations to the mean of each component to generate $\delta p(\mathbf{x})$. In 2D, the Green's function is $G_2(\mathbf{x}) = \frac{\hat{\mathbf{x}}}{2\pi |\mathbf{x}|}$. As the estimated flow may blow up when $\ell(\mathbf{x})$ is small, we also clipped the norm of the flow field so that no vector was more than 10 times the median norm. To visualize the estimated $\delta\hat{p}(\mathbf{x})$ against the true $\delta p(\mathbf{x})$, we computed $\mathbf{z}^i = \mathbf{x}^i + 0.0001 \delta \mathbf{x}^i$, took the difference between kernel density estimates with $\sigma=0.2$ and applied a median filter to the result to reduce the noise in regions with low probability and few samples. As can be seen in Fig.~1(c), this kernel density estimate closely matches the true perturbation.

To make this comparison more rigorous, we also evaluated the kernelized Stein discrepancy between the samples and the perturbed probability, a statistical test of goodness-of-fit which does not require knowledge of the true normalizing factor of the distribution to compare against \cite{chwialkowski2016kernel, liu2016kernelized}. Following \cite{liu2016kernelized}, we used an RBF kernel $k(\mathbf{x}, \mathbf{x}') = \mathrm{exp}(-\frac{1}{2\sigma^2} |\mathbf{x}-\mathbf{x}'|^2)$ with bandwidth given by the median of the sample distances. The U-statistic for the kernelized Stein discrepancy between samples $\mathbf{x}_i$ and a distribution with (possibly unnormalized) log probability $\ell(\mathbf{x})$ is then given by $\frac{1}{N(N-1)} \sum_{i \ne j} \nabla \ell(\mathbf{x}_i)^T k(\mathbf{x}_i, \mathbf{x}_j) \nabla \ell(\mathbf{x}_j) + \nabla \ell(\mathbf{x}_i)^T \nabla_{\mathbf{x}_j} k(\mathbf{x}_i, \mathbf{x}_j) + \nabla_{\mathbf{x}_i}k(\mathbf{x}_i, \mathbf{x}_j)^T \nabla \ell(\mathbf{x}_j) + \mathrm{Tr}\left(\nabla_{\mathbf{x}_i} \nabla_{\mathbf{x}_j} k(\mathbf{x}_i, \mathbf{x}_j) \right)$. Rather than bootstrapping samples and computing quantiles of the U-statistic, we simply report the mean and standard deviation of the U-statistic for 10 different values of $\delta p(\mathbf{x})$ on the same mixture of Gaussians example from Fig.~1. We computed this discrepancy both for the original data $\mathbf{x}^i$ and the data shifted by the flow $\mathbf{x}^i + \epsilon \mathbf{v}^i$ for different values of $\epsilon$. The results are given in Fig.~2. It can clearly be seen that, over a wide range of $\epsilon$, the discrepancy normalized by $\epsilon$ is roughly flat for the samples perturbed by the flow, while it is much larger at the original samples. As $\epsilon$ grows, eventually the perturbed distribution is no longer very close to the original distribution, and the perturbed particles are no longer a good approximation to samples from the distribution, but for moderate $\epsilon$ the integrable flow field clearly works well. We leave it to future work to expand this to more complex, higher-dimensional distributions.

\section{Discussion}

Integrable nonparametric flows are closely related to the ``space-warp coordinates" used in quantum Monte Carlo \cite{filippi2000correlated}. Space-warp coordinates are specifically for the case of computing forces by estimating the energy of a system where atomic nuclei are perturbed, and are computed based on a heuristic that electrons near the nucleus of one system should remain near that nucleus when perturbed. Integrable nonparametric flows, by contrast, are an exact method (in the limit $N\rightarrow\infty$) and can be applied to {\em any} problem where a probability distribution is perturbed a small amount.

In machine learning, the most closely related work is Stein Variational Gradient Descent (SVGD) \cite{liu2016stein} and the optimal Coulomb kernels method of Hochreiter and Obermayer \cite{hochreiter2005optimal}. In all methods, a small update is applied to a set of particles approximating a target distribution.
In our work, we aim to match an infinitesimally close target density {\em exactly} in a single step, while in both SVGD and optimal Coulomb kernels, the aim is to match a distant target density over several steps. In SVGD, a steepest descent direction is derived in the reproducing kernel Hilbert space (RKHS) over unit-norm functions, while in both our work and Hochreiter and Obermayer, the optimal step is constrained to be an integrable vector field. The critical difference between our work and SVGD is that, in SVGD, the update depends on the choice of RKHS. Different RKHS's induce different norms over distributions, leading to different steepest descent directions. In our work, since we aim to exactly match the target distribution in a single step, the descent direction in the space of distributions is independent of the choice of norm, which simplifies the math considerably. The derivation of our update is nearly the same as in Hochreiter and Obermayer (which we only became aware of in the course of writing the paper), the main difference being that they are interested in model selection for parametric unsupervised learning, while we are interested in accelerating the convergence of sampling when given unnormalized densities.

The relationship between integrable nonparametric flows and optimal transport \cite{peyre2019computational} also deserves attention. In both cases, the problem is to find a mapping between probability distributions, and investigating how these two problems are related in the limit of infinitesimal changes between the probability distributions is a promising direction for future work.

The major limitation to applying integrable nonparametric flows more broadly is the curse of dimensionality. As the Green's function $G_n$ falls off as $r^{n-1}$, and the estimation of the partition function becomes impractical in higher dimensions, we expect the method outlined here will struggle as the dimensionality grows. Much like in density estimation, prior knowledge could be built in that allows us to escape this curse in specialized cases, but we will leave it to future work to discover how.

\section*{Broader Impact}

The present work is primarily a proof-of-principle, and at present is only applicable to low-dimensional problems. If it could be scaled, it could be used to help make machine learning and quantum chemistry algorithms more efficient. The social impact of such an improvement would primarily be to accelerate the impact of whichever method is being made more efficient.

\begin{ack}
We would like to thank David Duvenaud for helpful discussions.
\end{ack}

\bibliographystyle{plain}
\bibliography{main}
\end{document}